\documentclass[10pt,twocolumn,letterpaper]{article}

\usepackage[pagenumbers]{cvpr} 

\RequirePackage{fix-cm}
\usepackage{siunitx}
\usepackage{graphicx}
\usepackage{svg}
\usepackage{makecell}
\usepackage{enumitem, amssymb, amsmath}
\usepackage{booktabs}
\usepackage{acronym}
\usepackage{multirow}
\newcommand{\eref}[1]{Eq.~(\ref{#1})}
\newcommand{\fref}[1]{Figure~\ref{#1}}
\newcommand{\sref}[1]{Section~\ref{#1}}
\newcommand{\tref}[1]{Table~\ref{#1}}
\acrodef{bev}[BEV]{bird's-eye view}
\acrodef{msa}[MSA]{multi-head self-attention}
\acrodef{slam}[SLAM]{simultaneous localization and mapping}
\definecolor{cvprblue}{rgb}{0.21,0.49,0.74}
\usepackage[pagebackref,breaklinks,colorlinks,allcolors=cvprblue]{hyperref}

\def\paperID{17047}
\def\confName{CVPR}
\def\confYear{2025}

\title{ForestLPR: LiDAR Place Recognition in {Forests} Attentioning Multiple \\BEV Density Images}

\author{Yanqing Shen$^{1,2}$, ~Turcan Tuna$^2$, ~Marco Hutter$^2$, ~Cesar Cadena$^2$, ~Nanning Zheng$^{1}$\thanks{Corresponding author}\\
$^1$Institute of Artificial Intelligence and Robotics, Xi'an Jiaotong University \\$^2$Robotic Systems Lab, ETH Zurich\\
{\tt\small {qing1159364090@stu,nnzheng@mail}.xjtu.edu.cn \quad {tutuna,mahutter,cesarc}@ethz.ch}
}

\date{March 2025}

\begin{document}
\maketitle

\begin{abstract}
Place recognition is essential to maintain global consistency in large-scale localization systems.
While research in urban environments has progressed significantly using LiDARs or cameras, applications in natural forest-like environments remain largely under-explored.
Furthermore, forests present particular challenges due to high self-similarity and substantial variations in vegetation growth over time.
In this work, we propose a robust LiDAR-based place recognition method for natural forests, ForestLPR.
We hypothesize that a set of cross-sectional images of the forest's geometry at different heights contains the information needed to recognize revisiting a place.
The cross-sectional images are represented by \ac{bev} density images of horizontal slices of the point cloud at different heights.
Our approach utilizes a visual transformer as the shared backbone to produce sets of local descriptors and introduces a multi-BEV interaction module to attend to information at different heights adaptively.
It is followed by an aggregation layer that produces a rotation-invariant place descriptor.
We evaluated the efficacy of our method extensively on real-world data from public benchmarks as well as robotic datasets and compared it against the state-of-the-art (SOTA) methods.
The results indicate that ForestLPR has consistently good performance on all evaluations and achieves an average increase of 7.38\% and 9.11\% on Recall@1 over the closest competitor on intra-sequence loop closure detection and inter-sequence re-localization, respectively, validating our hypothesis\footnote{\href{https://github.com/shenyanqing1105/ForestLPR-CVPR2025}{https://github.com/shenyanqing1105/ForestLPR-CVPR2025}}.
\end{abstract}
\acresetall
\section{Introduction}

Place recognition in natural forest-like environments is known to be challenging due to reduced salient features and high self-similarity. Forest localization is challenging even for humans, yet it is essential for robotic navigation, \ac{slam}, and augmented reality applications.
Under poor global positioning, place recognition can provide loop-closure constraints to mitigate the adverse effects of odometry drifts in mapping applications.

\begin{figure}[t]
    \centering
    \includegraphics[width=0.47\textwidth]{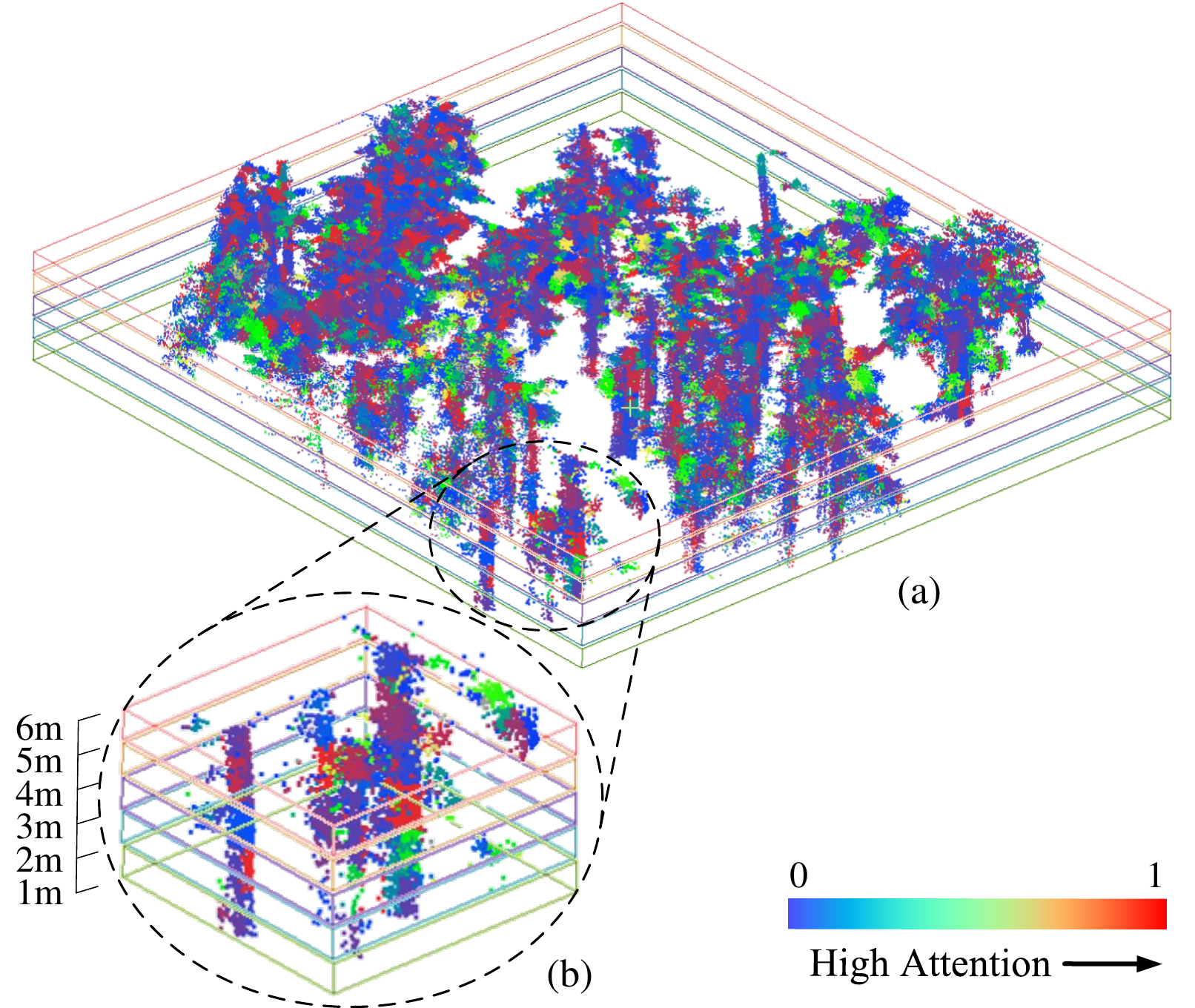}
    \vspace{-0.15cm}
    \caption{
    {(a) shows a part of the point cloud submap that is colored by projected attention maps (the zoomed-in part is shown in (b)).}
    It can be seen that after removing the ground and tree top, some canopy and bushes are still included in the cropped point cloud. 
    Our ForestLPR framework is able to utilize multiple BEV density images and a multi-BEV interaction module to achieve adaptive attention to different heights at each patch location.
    }
    \label{fig:motivation}
\end{figure}

Compared to the traditional registration-based methods~\cite{arun1987least}, feature-based methods have become widely adopted in LiDAR place recognition, as they directly extract global features from 3D point clouds~\cite{ma2022overlaptransformer,minkloc3dv2,transloc3d,logg3d} or 2D projections~\cite{scan, mapclosure,spi}. 
However, the primary application area of these methods is currently limited to urban environments.
Few works have extended the application to forest environments~\cite{wildplaces,oh2024evaluation}, adapting or fine-tuning the above learning-based models.
These works have shown that a domain gap between urban and forest environments is present, and attention from the community is needed.
Furthermore, the high variability within forest environments, such as tree species, age, and height, results in different point distributions. This makes it more difficult to generalize 3D models across diverse forests.

In this work, we build on two assumptions to deal with perceptual aliasing in natural forests: i) the \textit{unique spatial distribution} and ii) the \textit{discriminative features for different trees} can be found at different heights.
In order to aid the generalization of different terrains and tree species, we discard the ground and the canopy above a given height, as those are the most affected by seasonal changes. 
Specifically, we crop horizontal slices at different heights, above the ground, and under the canopy, to generate a bird's-eye view (BEV) density image of each slice.
Then, we propose a multi-BEV interaction module to combine the patch-aligned features of multiple BEV images, thereby implicitly identifying stable discriminative features and achieving adaptive attention to heights at each patch location.

We compare ForestLPR with SOTA methods on Wild-Places~\cite{wildplaces}, Botanic Dataset~\cite{liu2023botanicgarden}, and a dataset collected with a robotic platform (ANYmal Dataset).
To verify the generalization ability of our method, we use the training set of Wild-Places for training and then test the other two datasets without any fine-tuning.
Experimental results show that our method outperforms SOTA methods in multiple evaluations while achieving a better trade-off between performance and computational cost.
Furthermore, we conduct extensive ablation studies that validate the effectiveness of using multiple BEV density images and our proposed multi-BEV interaction module.

The contributions of this paper are as follows:
\begin{itemize}
    \item A multi-BEV interaction module that enables adaptive patch-level attention to different heights in forests, thereby mitigating information loss along the elevation axis. Our results demonstrate strong generalization across diverse forest conditions.
    \item An adapted Transformer-based feature extraction backbone that uses BEV density images from point clouds as input to preserve 2D geometry along the ground plane. 
    \item A novel LiDAR-based place recognition pipeline that incorporates essential preprocessing steps to address disturbances and seasonal vegetation variations in forests.
\end{itemize}
\section{Related Work}
LiDAR-based place description and recognition in outdoor environments is an active field of research, typically classified into two main categories: directly utilizing 3D point clouds as input and using projected 2D images as input.

\textbf{3D Point Cloud-based Methods.}
Recently, deep neural network based end-to-end methods~\cite{uy2018pointnetvlad,lcdnet,fan2022svt,zhou2021ndt,hui2021pyramid,locus,bevt,xia2021soe,zhang2019pcan,yang2024lisa} have been developed to derive scene representations from 3D point clouds.
PointNetVLAD~\cite{uy2018pointnetvlad} pioneered the use of an end-to-end trainable global descriptor for LiDAR place recognition, using PointNet~\cite{pointnet} and NetVLAD~\cite{netvlad}.
However, due to the lack of spatial distribution information on local features, it is hard to directly implement PointNet in large scenes.
To address this problem, LPD-Net~\cite{lpdnet} and EPC-Net~\cite{hui2022efficient} proposed adaptive local feature extraction and the graph-based neighborhood aggregation module.
SOE-Net~\cite{xia2021soe} explored the relationship between points and long-range context into local features.

More recently, many works~\cite{minkloc3d,egonn,transloc3d,lang2019pointpillars} have aimed to obtain descriptors from discretized representations of LiDAR scans.
Particularly, MinkLoc3D~\cite{minkloc3d} utilized sparse convolutions to capture useful point-level features.
Based on MinkLoc3D, MinkLoc3Dv2~\cite{minkloc3dv2} proposed to combine the loss function based on a smooth approximation of the Average Precision metric.
TransLoc3D~\cite{transloc3d} utilized adaptive receptive fields with a point-wise re-weighting scheme to handle objects of different sizes while suppressing noises and an external transformer to capture long-range feature dependencies.
Logg3D-Net~\cite{logg3d} proposed additional local consistency loss to guide the sparse convolutional network towards learning consistent local features across revisits.
Moreover, some approaches~\cite{segmatch2017,segmap1,segmap2018,semsegmap,locus,nardari2020place,nsm,esm,dsm} have been proposed based on segments.
However, conventional segmentation methods struggle in dense forests due to the intermingling of canopy and complex tree growth.

\textbf{2D Projection-based Methods.}
Several approaches converted 3D point clouds to 2D structures as intermediate representations and then used visual techniques to compute descriptors. 
2D representations include spherical-view range images~\cite{rangenet,steder2010robust, steder2011place, chen2021overlapnet}, BEV bins~\cite{scan}, and BEV images~\cite{luo2023bevplace,mapclosure,xu2023ring++}.
For example, an interest point extractor was proposed for range images~\cite{steder2010robust}.
OverlapNet~\cite{chen2021overlapnet} estimated the overlap and relative yaw angle between range images based on depth, normals, and intensity values.
To make the approach more accessible to generalize, OverlapTransformer~\cite{ma2022overlaptransformer} improved OverlapNet and created descriptors using only depth clues.
However, range images are not suitable for multi-frame registered submaps.

ScanContext~\cite{scan} captured height information from BEV images and integrated them into a 2D descriptor, and it was extended by using intensity~\cite{wang2020intensity} and semantics~\cite{li2021ssc}.
Still, these methods are sensitive to relative translation.
ScanContext++~\cite{kim2021scan} computed Cartesian elevation maps and polar elevation maps for translational invariance.
BEV image-based works treat projections as pixel-images~\cite{spi,luo2021bvmatch,jiang2023contour,yuan2023std}.
ContourContext~\cite{jiang2023contour} generated BEV projections at predefined heights, clustered the projected points, and computed the statistics from contours.
However, using an elevation map as the 2D representation is sensitive to the sensor viewpoint.
BVMatch~\cite{luo2021bvmatch} used density image representations to extract handcrafted FAST~\cite{rosten2006machine} key points but required training a bag-of-words model. Thus, it is not ideal for novel or unseen environments.
BEVPlace~\cite{luo2023bevplace} used group convolution~\cite{cohen2016group} to extract local features and generated global features through NetVLAD~\cite{netvlad}.
MapClosures~\cite{mapclosure} proposed to extract ORB features of BEV density images.

We use BEV density images from point clouds because suitable BEV projections can preserve 2D geometry along the ground plane, which is crucial for place recognition in forests.
Moreover, our solution uses multiple BEV density images generated from slices of different heights for each point cloud. It achieves adaptive attention to various heights, avoiding information loss on the elevation axis.
\section{Methodology}

\subsection{Pre-processing of Point Cloud}
\label{sec:pre}
The ground of natural forests can vastly vary in height, slope, and appearance. 
In a similar manner, the forest canopy is primarily affected by weather (e.g., wind or seasons).
Thus, we pre-process the point cloud to exclude this variability by removing the height offset and removing the ground and the tree tops above a certain height.

\begin{figure}[t]
    \centering
    \hspace{-4mm}
\subfloat[Ground segmentation]{
    \label{fig:ground}
  \includegraphics[width=0.48\columnwidth]{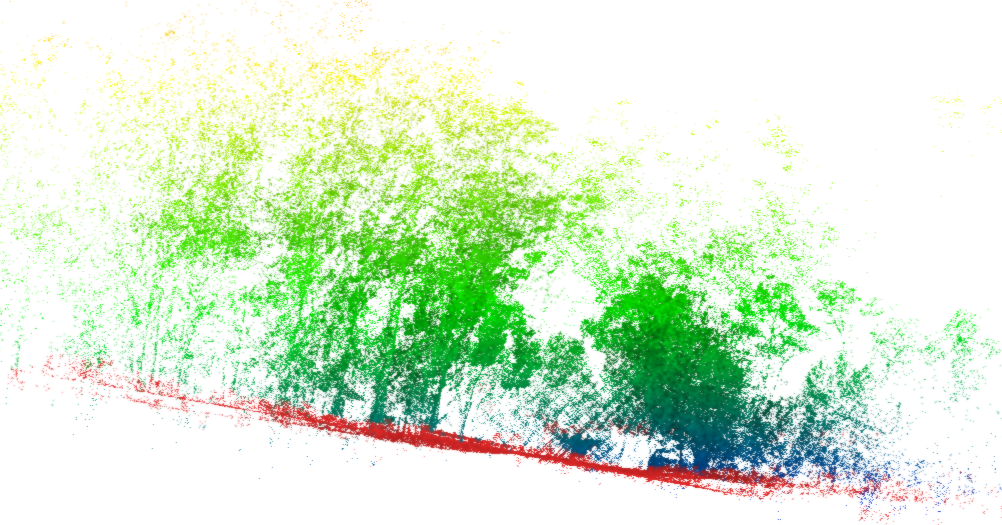}}
\subfloat[Height offset removal]{
    \label{fig:height}
  \includegraphics[width=0.48\columnwidth]{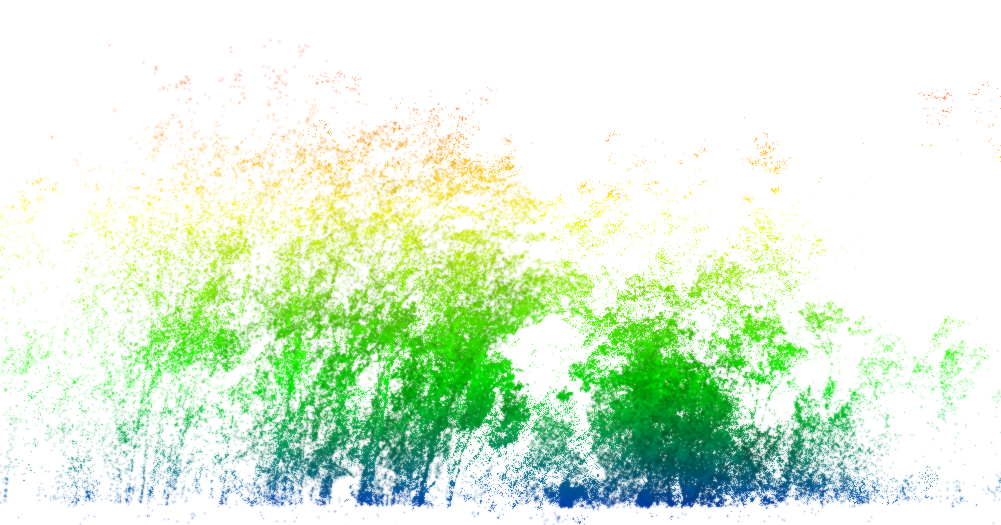}}
    \caption{Results of ground segmentation and height offset removal. In (a), the ground points are marked in red. The ground segmentation result is used to remove the height offset of non-ground points, as shown in (b). Point clouds are colored by height.}
\end{figure}

\textit{1) Ground Segmentation and Height Offset Removal.}
We first minimally denoise the cloud and apply ground segmentation~\cite{zhang2016easy} to distinguish ground points from non-ground points.
An example result of the ground segmentation is given in \fref{fig:ground}.

Given that tree height refers to the height of points from the ground, it is necessary to discount the ground height under a particular tree~\cite{lucatree}.
For each non-ground point $\mathbf{p}$, we search its \textit{neighbor ground points}, $\mathbf{p}_g$, in the ground point set.
The L2 distance over the x-y coordinates of $\mathbf{p}$ and $\mathbf{p}_g$, $d(\mathbf{p}, \mathbf{p}_g)$, is less than a radius $R$.
Due to the blind spots in datasets (Figure 1 in the supplementary material), some non-ground points do not have neighbor ground points in the above calculation. 
For these points, the radius $R$ is gradually increased by $\bigtriangleup R$ meter until $\mathbf{p}$ has neighbor ground points or the radius reaches a maximum radius, $R_{max}$.
If neither condition is met, this non-ground point is removed.

For each remaining non-ground point $\mathbf{p}$, the average elevation of its neighbor ground points is subtracted from its original elevation $z^p$, so the final height is
\begin{equation}
h(\mathbf{p}) = z^p - \frac{\sum d\left(\mathbf{p}, \mathbf{p}_g\right)^{-2} z^g}{\sum d\left(\mathbf{p}, \mathbf{p}_g\right)^{-2}} ,\\
\end{equation}
where $z^g$ refers to the height of $\mathbf{p}_g$.
An example of height normalization is given in \fref{fig:height}, converting the non-ground point cloud to a flat terrain.

\textit{2) Ground and Tree Top Removal.}
Now that all the root collars \footnote{The point at or just above the ground level of a tree trunk.} are at the same height, we remove as many highly variable elements as possible.
We give the margin of \SI{1}{\meter} to account for terrain variations like grass, leaves, snow, and fallen branches.
Similarly, we remove points above \SI{6}{\meter} to reduce inconsistencies caused by seasonal canopy changes, as well as sparser data and occlusions due to scanning from a ground platform.

\subsection{BEV Density Images from Point Cloud}
\label{sec:bev}
To ensure robustness across different configurations of LiDAR, recent approaches~\cite{luo2021bvmatch,luo2023bevplace,mapclosure} use BEV density images and have shown better performance in urban environments than other methods.
Thus, we compute BEV density images from pre-processed point clouds.

\textit{1) 2D BEV Projection.} Each pre-processed point cloud is projected onto the ground plane and discretized into a 2D grid through a Cartesian BEV projection, of which the resolution is $res$ for each cell.

\begin{figure}[t]
    \centering
    \includegraphics[width=\linewidth]{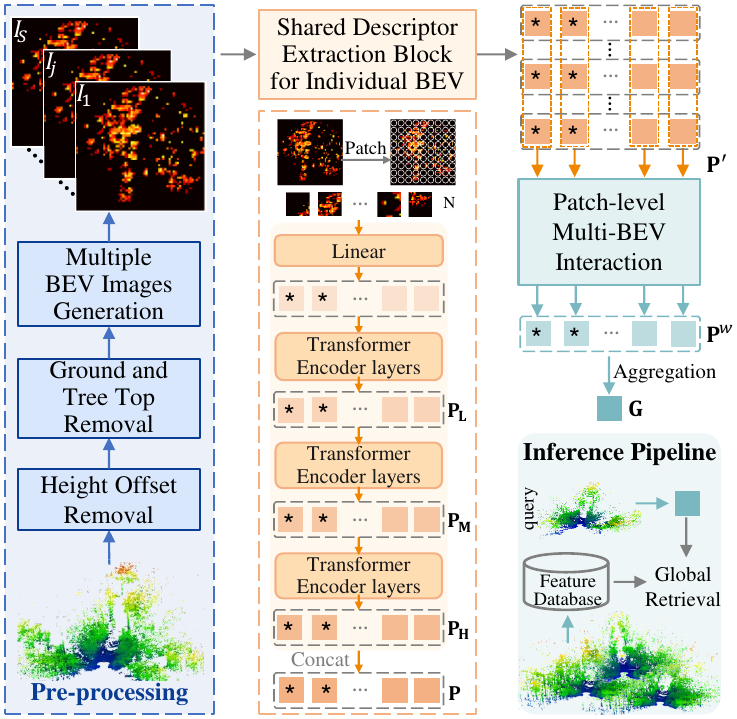}
    \caption{An overview of the proposed framework. After pre-processing, multiple BEV density images are generated from a point cloud. Each BEV image can be processed separately by the shared descriptor extraction block, combining features from multiple transformer layers. The local features are then fed into the path-level multi-BEV interaction module to highlight the discriminative features. Global descriptors are obtained through an aggregation layer. The query and database share the same pipeline to extract features during training and testing.}
    \label{fig:pipeline}
\end{figure}
 
{The density value $V$ of pixel $(u,v)$ is,}
\begin{equation}
    V(u,v) = log(V'(u,v) +1),
\end{equation}
where $V'(u,v)$ is the number of points in the cell.
Such $log$ computation avoids the dominance of dense areas and preserves sparse areas during the learning process.
Then, the density image is defined as,
\begin{equation}
   I(u,v) =  \frac{V(u, v)-V_{\min }}{V_{\max }-V_{\min }},
\end{equation}
\begin{equation}
    V_{\max }=\max _{u, v} V(u, v) ; V_{\min }=\min _{u, v} V(u, v) .
\end{equation}
Such a projection preserves 2D spatial distribution.

\textit{2) Multi-BEV Representation.}
Single BEV images (create only one BEV image from the point cloud) keep the rigid structures on the 2D plane but ignore the point distribution along the z-axis, leading to information loss in the elevation direction.
To address this limitation, we cut the pre-processed point cloud into $S$ slices with a height interval of $\bigtriangleup h$. 
This segmentation represents a multi-layered distribution of structures.

\subsection{Descriptor Extraction Block}
\label{sec:block}
Here, we describe how the local features are generated from individual BEV density images, as shown in the orange dotted box in \fref{fig:pipeline}.
Given that the \ac{msa} operation in vision transformers~\cite{dosovitskiy2020image,deit} can aggregate global contextual information, it is suitable for extracting the overall spatial distribution features.

In this work, we add an extra linear layer before the encoder of data-efficient image transformer (DeiT)~\cite{deit} to match the input size.
Then a BEV input tensor ${I} \in \mathbb{R}^{H\times W \times 1}$ is first divided into many small patches of size $p \times p$ and converted into a set of tokens $T \in \mathbb{R}^{N \times C}$ by linear projection.
Here ($H,W$) is the resolution of the resized image, $N, C$ denotes the number and dimension of embedding tokens, with $N=H W/p^2$. 
In addition to $N$ patch tokens, DeiT adds learnable {\tt [class]} and {\tt [distillation]} tokens, which can be used in global descriptors.

It is observed that there are some differences in the scales of structures captured by different Transformer layers~\cite{transvpr}. 
To integrate information across multiple layers, we select three groups of output patch tokens from the low-level, mid-level, and high-level layers, denoted as $\{ \mathbf{P_L}, \mathbf{P_M}, \mathbf{P_H} \} \in \mathbb{R}^{(N+2)\times C}$. A group of multi-layer patch tokens $\mathbf{P} \in \mathbb{R}^{(N+2)\times 3C}$ are firstly composed by concatenating these three groups of tokens along the channel.

\subsection{{Multi-BEV} Interaction Module}
\label{sec:whole}
Given that some canopy and bushes are still included in the pre-processed point cloud, we propose to perform explicit feature interaction in the z-axis to improve the generality.
As shown in \fref{fig:pipeline}, for $S$-slice input, each slice, $I_j$, is processed by the shared extraction block separately, and the set of local features is $\mathbf{P'} \in \mathbb{R}^{ (N+2) \times S \times 3C}$.
Then, we design a patch-level multi-BEV interaction module on features from different heights. 

Specifically, we generate weight vectors for each patch to achieve adaptive attention to heights.
Inspired by channel-wise attention~\cite{bastidas2019channel}, there are two considerations in designing the weighting layer: \textit{1)} Use a fixed dimension of input to make the layer compatible with different sizes of $S$ for various applications, and \textit{2)} contain the patch-level context and capture the relative relationship in input.

For token $i$, we first generate $\Delta \mathbf{P'}_i \in \mathbb{R}^{S \times 3C}$,
\begin{equation}
    \Delta \mathbf{P'}_i = \mathbf{P'}_i - \underset{j}{\mathrm{Mean}} ( \mathbf{P'}_{i, j}),
\end{equation}
where $\Delta \mathbf{P'}_i$ represents the relative feature value.
The patch-level weight vector can be generated by
\begin{equation}
\label{eq:weight}
    \textbf{w}_i = \mathrm{SoftMax}(\Delta \mathbf{P'}_i * \textbf{W}_a) \in \mathbb{R}^{S \times 1},
\end{equation}
\begin{equation}
    \mathbf{P}^w_i =  \mathbf{w}_i^T * \mathbf{P'}_i \in \mathbb{R}^{3C},
\end{equation}
where $\mathbf{W}_a \in \mathbb{R}^{3C \times 1}$, and $\mathbf{P}^w \in \mathbb{R}^{(N+2) \times 3C}$ consists of weighted tokens.

\begin{figure*}[t]
    \centering
    \includegraphics[width=\linewidth]{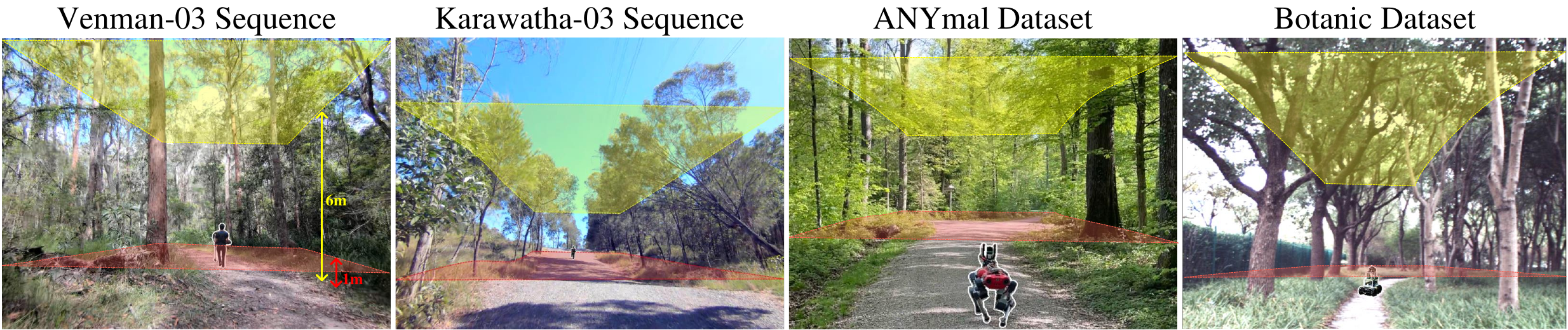}
    \vspace{-0.5cm}
    \caption{Environment and equipment visualization of different datasets, showing diversity and richness. We use red planes to illustrate locations where the height is \SI{1}{\meter} and yellow planes to indicate locations at a height of \SI{6}{\meter}. It can be shown that although most of the leaves and bushes are removed, some are still included in the horizontal slices rather than having a perfectly clear appearance, which brings great challenges to methods and motivates us to use multiple BEV density images in this work.}
    \label{fig:environ}
\end{figure*}

\paragraph{Global Aggregation Layer.}
A lightweight pooling layer, GeM~\cite{gem}, is applied to obtain yaw-invariant global features from patch-level local descriptors. Then, two global tokens are concatenated:
\begin{equation}
\label{eq:global1}
    \mathbf{G}^* = \mathrm{Concat}(\mathbf{P}^w[0,:],\mathbf{P}^w[1,:],\mathrm{GeM} (\mathbf{P}^w[2:,:])),
\end{equation}
where $\mathbf{P}^w[0,:]$ and $\mathbf{P}^w[1,:]$ refer to weighted {\tt [class]} and {\tt [distillation]} tokens. And then the final global feature $\mathbf{G}$ is obtained by post-processing $\mathbf{G}^* \in \mathbb{R}^{9C}$:
\begin{equation}
\label{eq:global2}
    \mathbf{G} = \mathrm{L2Norm}( \mathrm{L2Norm} (\mathbf{G}^*) * \mathbf{W}_g )),
\end{equation}
where $\mathbf{W}_g \in \mathbb{R}^{9C \times D}$ is used for dimension reduction, $D$ is the dimension of $\mathbf{G}$, and $\mathrm{L2Norm}(\mathbf{x}) = \frac{\mathbf{x}}{\left \| \mathbf{x} \right \| }$.

Finally, we modify the DeiT by removing the classification layer and adding the proposed global aggregation layer.

\subsection{Network Training}
\label{sec:train}
At first, we generate single BEV images from the pre-processed point cloud, and global features are computed by replacing $\mathbf{P}^w$ in \eref{eq:global1} with $\mathbf{P}$. Then, the extraction block is fine-tuned on the training set of the Wild Places dataset.

After block initialization, the whole ForestLPR with multi-BEV interaction module can be further finetuned in an end-to-end fashion on the training set of Wild-Places.

In these two training steps, the commonly used triplet loss~\cite{schroff2015facenet} is adopted to be the training objective, defined as:

\begin{footnotesize}
\centering
\begin{equation}
    \mathcal{L}(\mathbf{G^q},\mathbf{G^p},\mathbf{G^n}) = \mathrm{max}(d(\mathbf{G^q},\mathbf{G^p})-d(\mathbf{G^q},\mathbf{G^n})+m,0),
\end{equation}
\end{footnotesize}
where $\mathbf{G^q}$, $\mathbf{G^p}$, and $\mathbf{G^n}$ are global features of query, positive samples, and negative samples. The margin $m$ is a constant hyperparameter. $d(\cdot)$ refers to the {Cosine distance.}

\paragraph{Overlap-based Positive Sample Mining.}
Two samples are usually considered a true positive match if their distance falls below a threshold~\cite{scan, jiang2023contour}. 
It fails in object occlusion, limited field of view, and significant orientation variations~\cite{leyva2023data,mapclosure}.
Due to blind spots in our benchmarks (shown in the supplementary material), it is necessary to define more realistic ground truth for training and evaluation.

We utilize volumetric overlap and use Octree to find the overlapped region.
For each query and its potential positive, we use the ground truth poses to align them and voxelize them as $\mathcal{V}_p$ and $\mathcal{V}_q$.
The overlap can be computed as:
\begin{equation}
\label{eq:overlap}
    o\left(\mathcal{V}_q, \mathcal{V}_p\right)=\frac{\left|\mathcal{V}_q \cap \mathcal{V}_p\right|}{\left|\mathcal{V}_q \cup \mathcal{V}_p\right|},
\end{equation}
where $o\left(\mathcal{V}_q, \mathcal{V}_p\right)$ replaces the distance measure and serves as the reference for mining positives, where $o > 0.9$.
\section{Implementation Details}

\subsection{Datasets}

\textbf{Wild-Places~\cite{wildplaces}.}
This is a challenging large-scale dataset for LiDAR place recognition in forests. It contains 8 LiDAR sequences in Venman and Karawatha and was collected using a handheld sensor payload over fourteen months.
There are different types of evaluations involving reverse revisits: \textit{validation in training, intra-sequence loop closure detection, and inter-sequence re-localization.}
The point clouds are sampled submaps with a diameter of \SI{60}{\meter}.
Following the standard dataset division of Wild-Places, we use the training/validation set of Sequences 01 and 02 for training/validation and perform intra-/inter-sequence evaluation on the defined sequences~\cite{wildplaces}.

\textbf{ANYmal Dataset.} 
We collected a dataset in forests from a low perspective using the quadrupedal robot ANYmal\cite{ANYmal,ANYmal1}, equipped with a Velodyne VLP-16 scanner. 
For submap generation, we refer the reader to Wild-Places~\cite{wildplaces}.

\begin{table*}[t]
  \centering
   \caption{Benchmarking results of Intra-sequence evaluation on several standard datasets. V-03, V-04, K-03, and K-04 are subsets of Wild-Places. \textit{Bo.} refers to Botanic dataset. Point-based methods are evaluated on raw point clouds. PointPillar and BEV-based methods are marked with *, showing the fine-tuning results on pre-processed point clouds and density images for fair comparison and their better performance. For TransLoc3D\dag, we select more suitable hyper-parameters than those in Wild-Places.}
  \vspace{-0.2cm}
  \tabcolsep=0.15cm
  \label{tab:compare_SOTA_intra}
   \scalebox{0.85} {
  \begin{tabular}{lcccccccccccccccc}
  \toprule
{\multirow{2}{*}{Method}}& \multicolumn{2}{c}{V-03} & \multicolumn{2}{c}{V-04} & \multicolumn{2}{c}{K-03} & \multicolumn{2}{c}{K-04} & \multicolumn{2}{c}{ANYmal} & \multicolumn{2}{c}{Bo.1005-03} & \multicolumn{2}{c}{Bo.1005-06} & \multicolumn{2}{c}{Mean}\\
\cmidrule(lr){2-3} \cmidrule(lr){4-5} \cmidrule(lr){6-7} \cmidrule(lr){8-9} \cmidrule(lr){10-11} \cmidrule(lr){12-13} \cmidrule(lr){14-15} \cmidrule(lr){16-17} 
& F1 & R@1 & F1& R@1 & F1 & R@1 & F1 & R@1 & F1 & R@1 & F1 & R@1 & F1 & R@1 &F1 & R@1 \\
\midrule
PointPillar*~\cite{lang2019pointpillars}& 28.15&37.48&73.62&62.89&41.83&39.02&77.13&70.04&50.93&41.68&65.94&70.19&70.66&62.85 &58.32&54.88\\
TransLoc3D\dag~\cite{transloc3d} & 22.97 & 35.50& 75.00& 64.03& 43.44& 38.95 &75.78 &75.53 & 49.54& 38.26 &71.18 &76.89 &76.79 & 68.66&59.24&56.69\\
MinkLoc3Dv2~\cite{minkloc3dv2}& 49.78& 49.84& \textbf{82.19} & 71.61&51.33&50.97&80.00&71.18&72.89&66.83&29.98&43.11&31.11&34.33& 56.75&55.41\\
LoGG3D-Net~\cite{logg3d}& \underline{53.94}&\underline{62.40}& \underline{80.42}&72.47&\underline{64.30}&\underline{64.05}&\textbf{84.54}&\underline{80.26}&66.70& 56.42&\textbf{78.26}&\underline{79.78} & \underline{77.88} & \underline{82.09}&\underline{72.29}&\underline{71.07}\\ 
Scan-Context*~\cite{scan}&37.66&54.80&64.49&\underline{75.99}&34.33&50.23&66.85&63.48&\underline{73.44}&\underline{67.81}&34.34& 49.33&60.55& 76.12&53.09&62.54\\ 
BEVPlace*~\cite{luo2023bevplace}& 6.79&24.04&43.67&63.50&32.87&{40.62}&60.82&\textbf{81.74}&62.14&53.35&68.69&75.56&76.27&\textbf{85.08}&50.18&60.56\\ 
MapClosures*~\cite{mapclosure}& 38.47&23.82&49.95&25.61&21.35&11.95&70.62&54.59&71.54&55.77&{76.08}&75.56&73.59&58.21&57.37&43.64\\
\midrule
Ours &\textbf{64.15}&\textbf{76.53}&78.62&\textbf{82.33}&\textbf{65.01}&\textbf{74.89}&\underline{81.97}&{76.73}&\textbf{81.45}&\textbf{71.87}&\underline{78.21}&\textbf{84.81}&\textbf{78.82}& {82.00}&\textbf{75.46}&\textbf{78.45}\\
\bottomrule
\end{tabular}}
\end{table*}

\textbf{Botanic Dataset.}
Similarly, we repurpose the \textit{BotanicGarden} dataset~\cite{liu2023botanicgarden}, which is collected by Velodyne VLP-16 LiDAR.
Its environment is not dense forests, allowing for more diverse testing across different scenarios.

\textbf{Testing Generalization.}
We perform intra-sequence evaluation on ANYmal and Botanic datasets without additional fine-tuning.
\fref{fig:environ} provides the visualization of benchmarks, showing high diversity. For Wild-Places, the trees are tall and dense, with few low bushes. For ANYmal, the trees are of medium height, with a lot of low vegetation. For Botanic, the trees are short and sparse.

\subsection{Evaluation Metrics}
Following the setting in~\cite{wildplaces}, we consider a query successfully localized if the retrieved candidate is within \SI{3}{\meter}.
In addition, we use Recall@1 (R@1) and maximum F1 score (F1) as the metric for intra-sequence evaluation on all datasets, and R@1 is used for the inter-sequence assessment on Wild-Places in the paper.
More details and comprehensive results are given in the supplementary material.

\subsection{Training Details}

\textbf{Model Setting.}
ForestLPR is implemented using PyTorch~\cite{pytorch}.
We use DeiT-S-384~\cite{deit} as the backbone with $C$ = 384.
The patch size on the original image is set to $16 \times 16$.
Our extraction block contains 12 transformer encoder layers and 6 attention heads for each layer.
Without loss of generality, output tokens from the second, seventh, and twelfth transformer layers are selected as $\mathbf{P_L}$, $\mathbf{P_M}$, and $\mathbf{P_H}$.
The dimension $D$ of final global features is set to 1024.

\textbf{Training Strategy.}
We load the DeiT-S-384 model parameters pre-trained on ImageNet-1k\footnote{\href{https://dl.fbaipublicfiles.com/deit/deit_small_distilled_patch16_224-649709d9.pth}{deit\_small\_distilled\_patch16\_224-649709d9.pth}} and randomly initialize the parameters of the newly added layers.
After being fine-tuned on Wild-Places, the same model is used for all evaluation experiments.
$H \times W = 480 \times 480$ for training and evaluation.
We apply data augmentation by randomly rotating the pre-processed point clouds around the z-axis (perpendicular to the ground) within the $(\pi, \pi)$ interval.

\textbf{Hyper-parameters.}
For ground filtering, we follow the standard settings of CSF\footnote{\href{https://github.com/jianboqi/CSF}{https://github.com/jianboqi/CSF}}, except that $cloth\_resolution$ was changed to 0.3.
For height offset removal, $R=\SI{3}{\meter}$, $\bigtriangleup R=\SI{1}{\meter}$, $R_{max}=\SI{10}{\meter}$.
For cropping point cloud, {$S=5$} and $\bigtriangleup h=\SI{1}{\meter}$,
where \SI{1}-\SI{6}{\meter} is used for all datasets in the experiments.
The submaps are \SI{60}{\meter} in size, and $res=\SI{0.5}{\meter}$ for BEV image generation. More details are given in the supplementary material.

\section{Experimental Results}

We first show the results of our approach in comparison with existing state-of-the-art methods and then analyze the effectiveness of the proposed components by ablation studies.
More results are given in the supplementary material.

\subsection{Comparisons with State-of-the-Art Methods}
\label{sec:sota}
\textbf{Baselines.}
We compare ForestLPR with previous state-of-the-art baselines, including point cloud-based methods: PointPillars~\cite{lang2019pointpillars},  TransLoc3D~\cite{transloc3d}, MinkLoc3Dv2~\cite{minkloc3dv2}, and LoGG3D-Net~\cite{logg3d}, and BEV-based methods: ScanContext~\cite{scan}, BEVPlace~\cite{luo2023bevplace}, and MapClosures~\cite{mapclosure}.
To give a fair comparison in \tref{tab:compare_SOTA_intra} and \tref{tab:compare_SOTA_inter}, we use BEV density images generated from the pre-processed point cloud as the input of BEV-based methods and PointPillar, as well as re-train PointPillar and BEVPlace.
We re-run the pre-trained models of TransLoc3D, MinkLoc3Dv2, and LoGG3D-Net, which are provided by Wild-places\footnote{\href{https://csiro-robotics.github.io/Wild-Places/}{https://csiro-robotics.github.io/Wild-Places/}}. For TransLoc3D, we select more suitable hyper-parameters than those in Wild-Places~\cite{wildplaces}. (For more details and results, refer to the supplementary material.)

\textbf{Intra-run Loop Closure Detection on all datasets.}
In this case, each submap in a sequence acts as a query to match against previously seen point clouds in the same sequence, akin to the loop-closure detection problem.
According to the average ranking of methods on all datasets, LoGG3D-Net achieves the best performance among the baselines.
ForestLPR shows consistently good performance, see \tref{tab:compare_SOTA_intra}, achieving an average increase of 7.38\% on Recall@1 over SOTA performance on all datasets.

Across four evaluations of Wild-Places, ForestLPR shows an average increase of F1 and R@1 scores compared to LoGG3D-Net, 1.64\% and 7.83\%, respectively.
Judging from the performance shown in the \tref{tab:compare_SOTA_intra}, V-03 and K-03 pose a greater challenge, which is because they contain a large number of reverse revisits, and in dense forests, a large portion of the environment has non-overlapping regions due to occlusion.
However, ForestLPR outperforms other methods in these scenarios because it adaptively focuses on scene-specific features. 
For V-04 and K-04, our performance is comparable to SOTA, with only a slight decrease.
For example, BEVPlace achieves SOTA on R@1 of K-04 under uniformed tree growth at the expense of the highest computational load of all tested methods.

Furthermore, ForestLPR performs well on ANYmal and Botanic datasets without fine-tuning, showing its generalization on forest scenes with various tree growth conditions.
For ANYmal and Botanic datasets, differences in data collection from Wild-Places lead to different point distributions, which results in performance degradation when the method exhibits sensitivity to point distribution variations.
Compared with all methods on the ANYmal dataset, ForestLPR outperforms the SOTA method (Scan-Context) by 8.01\% and 4.06\% on F1 and R@1, respectively.
For the Botanic dataset, ForestLPR achieves an average improvement over LoGG3D-Net of 0.45\% and 2.47\% on F1 and R@1, respectively.

\begin{table}[t]
  \centering
   \caption{R@1 performance of Inter-sequence evaluations on Wild-Places. Inter-V/K refers to Inter-sequence evaluations, and Validation refers to the testing dataset in training. Point-based methods are evaluated on raw point clouds. PointPillar and BEV-based methods are marked with \textbf{*}, showing the fine-tuning results on pre-processed point clouds and density images for fair comparison and their better performance. For TransLoc3D\dag, we use more suitable hyper-parameters than those in Wild-Places.}
  \vspace{-0.2cm}
  \label{tab:compare_SOTA_inter}
\scalebox{0.9}{
  \begin{tabular}{lcccc}
  \toprule
{{Method}}& {Validation} & {Inter-V} & {Inter-K} & Mean\\
\midrule
PointPillar*~\cite{lang2019pointpillars}&35.49&60.34&52.19 & 49.34\\
TransLoc3D\dag~\cite{transloc3d} & 37.75&62.85&54.32&51.64\\
MinkLoc3Dv2~\cite{minkloc3dv2}& 51.40&75.77&67.82&65.00\\
LoGG3D-Net~\cite{logg3d}& \underline{54.34}&\textbf{79.84} & \underline{74.67}&\underline{69.62}\\
Scan-Context*~\cite{scan}&-&57.23&52.81&-\\
BEVPlace*~\cite{luo2023bevplace}& 26.25&51.91&41.40&39.85\\
MapClosures*~\cite{mapclosure}& -&34.31&19.35&-\\
\midrule
Ours&\textbf{80.03}& \underline{77.14}&\textbf{79.02} & \textbf{78.73}\\ 
\bottomrule
\end{tabular}}
\end{table}

\textbf{Inter-run Re-localization on Wild-Places.}
For inter-sequence evaluation, like a re-localization problem, we evaluate each sequence as queries against the 3 other sequences as databases in the same environment, giving 12 evaluations per environment.
Validation in~\tref{tab:compare_SOTA_inter} refers to the testing subset over all sequences, used for evaluating the training progress.
According to the average ranking of methods on all datasets, LoGG3D-Net and MinkLoc3Dv2 achieve top 2 performance among the baselines.

In~\tref{tab:compare_SOTA_inter}, most methods perform poorly.
That's because the sequences are collected at different times, and there are variations in tree growth. 
In contrast, ForestLPR, MinkLoc3Dv2, and LoGG3D-Net can achieve relatively high performance.
ForestLPR achieves an average increase of 9.11\% and 13.73\% on Recall@1 over the top 2 performance on three inter-sequence evaluations, respectively.
Specifically, since the leaves and growth of trees in the Karawatha dataset vary more across seasons than in Venman, the better performance of ForestLPR on inter-Karawatha supports its advantages in highly variable scenarios.

\begin{table}[t]
  \centering
\caption{Comparison of computational cost regarding feature dimension, latency, and memory footprint. All methods are measured on V-04 using the same CPU and GPU (RTX 3090). The batch size of learning-based models is 1.}
\vspace{-0.2cm}
\label{tab:latency_memory}
\scalebox{0.78}{
\begin{tabular}{lcccc}
\toprule
\multirow{2}{*}{Method} & \multicolumn{2}{c}{Latency per Query (ms) $\downarrow$}
&Memory $\downarrow$ & \multirow{2}{*}{Dim $\downarrow$}\\
\cmidrule(l{0pt}r{0pt}){2-3}
&Extraction&Retrieval & (MB)\\
\midrule
PointPillar~\cite{lang2019pointpillars}& 32.6&\underline{20.2}&840&1024\\
TransLoc3D~\cite{transloc3d} &25.09& \textbf{3.2} &1460&256\\
MinkLoc3Dv2~\cite{minkloc3dv2}& 24.74 & \textbf{3.2}&1382&256\\
LoGG3D-Net~\cite{logg3d}&64.26&\underline{20.2}&2046&1024\\
Scan-Context~\cite{scan}  &\underline{1.92}&384.2&-&1200\\
BEVPlace~\cite{luo2023bevplace} &116.05&99.25&634& 8192\\ 
MapClosures~\cite{mapclosure}&\textbf{0.36}&1400&-&-\\
\midrule
ForestLPR (ours) & 16.9 & \underline{20.2} &538 &1024\\ 
\bottomrule
\end{tabular}}
\end{table}

\textbf{Computational Cost.}
\tref{tab:latency_memory} presents the computational cost of methods, with the implementation aligned with those in \tref{tab:compare_SOTA_intra} and \tref{tab:compare_SOTA_inter}. All the methods are measured on Wild-Places V-04 (5,739 database submaps).
For BEV-based methods, pre-processing time is fixed (30 ms), including pre-processing point cloud and BEV generation.

Except for Scan-Context and MapClosures, which require additional alignment, all other methods use global feature vectors for retrieval.
With a descriptor size of 1024 and a retrieval time of 20.2 ms, ForestLPR has the potential to scale well to real-time robotics applications.

\subsection{Ablation Studies}
\label{sec:ablation}

In this section, we validate the proposed modules.

\textbf{Multiple Density Images.}
Without using multiple BEV images, the whole framework degenerates into generating one image from a pre-processed point cloud (\textit{single BEV}).
In \tref{tab:ablation_interaction}, using multiple BEV images can perform better in all tests.
In our experiments, tree growth in forests varies, and the pre-processed point cloud may still include some bushes and canopy rather than having a perfectly clear appearance.
So the features of trees can vary at different heights, and there are occlusions and perceptual aliasing in single BEV images, which contrasts with urban scenes.

\begin{table}[t]
  \centering
\caption{Ablation experiments on multiple BEV images and multi-BEV interaction module. R@1 performance is shown here.}
\vspace{-0.2cm}
\label{tab:ablation_interaction}
\scalebox{0.9}{
\begin{tabular}{lccccc}
\toprule
{Method}&{V-03} & {V-04} & {K-03} & {K-04}& ANYmal\\
\midrule
single BEV & 53.70 & 64.46&54.72&70.03&58.51\\
concatenation& 68.17&64.25&55.84&71.04&\underline{63.26}\\ 
w/o weighting& 59.68&\underline{76.87}& 63.00& 71.81&58.68\\ 
w/o interaction&\underline{73.76}&76.03&\underline{67.22}&\underline{72.23} &62.62\\ 
\midrule
Ours & \textbf{76.53} & \textbf{82.33} & \textbf{74.89}&\textbf{76.73}&\textbf{71.87}\\
\bottomrule
\end{tabular}}
\end{table}

\textbf{Multi-BEV Interaction Module.}
We conduct ablation studies on the proposed multi-BEV interaction module, mainly focusing on separate feature extraction, multi-BEV aggregation, and interaction-based weighting.
Three degraded designs are as follows: 1) \textit{Concatenation input}: multiple density images are concatenated and processed as a single unit to extract features through the modified backbone, 2) \textit{Aggregation without weighting}: use max pooling to fuse multiple local descriptors for each patch, 3) \textit{Weighting without interaction}: original local descriptors are used to generate weights, meaning there is no interaction between descriptors of each patch when generating weights. After re-training, the results are shown in \tref{tab:ablation_interaction}.

For \textit{concatenation}, we adjust the linear layer before the backbone to match the input size and obtain global features by placing the aggregation layer on top of the output local features.
In this case, performance drops by an average of 13.96\%. This suggests that feature-level interaction in the backbone is insufficient, and it is essential to pay explicit attention to each slice based on separate feature extraction.
For \textit{w/o weighting}, we use max pooling to aggregate local features for each patch.
It considers different heights independently but is less effective than weighting, resulting in an average performance drop of 10.46\%. 
For \textit{w/o interaction}, $\Delta \mathbf{P}_j(i)$ is replaced by $\mathbf{P}_j(i)$ in \eref{eq:weight}.
The average performance decrease of 6.10\% indicates that our interaction module captures the implicit relationship between different heights more effectively than using the absolute value of features, thereby enhancing discriminative features.



\begin{figure}[t]
    \centering
    \hspace{-4mm}
\subfloat[ANYmal dataset]{
    \label{fig:any}
  \includegraphics[width=0.48\columnwidth]{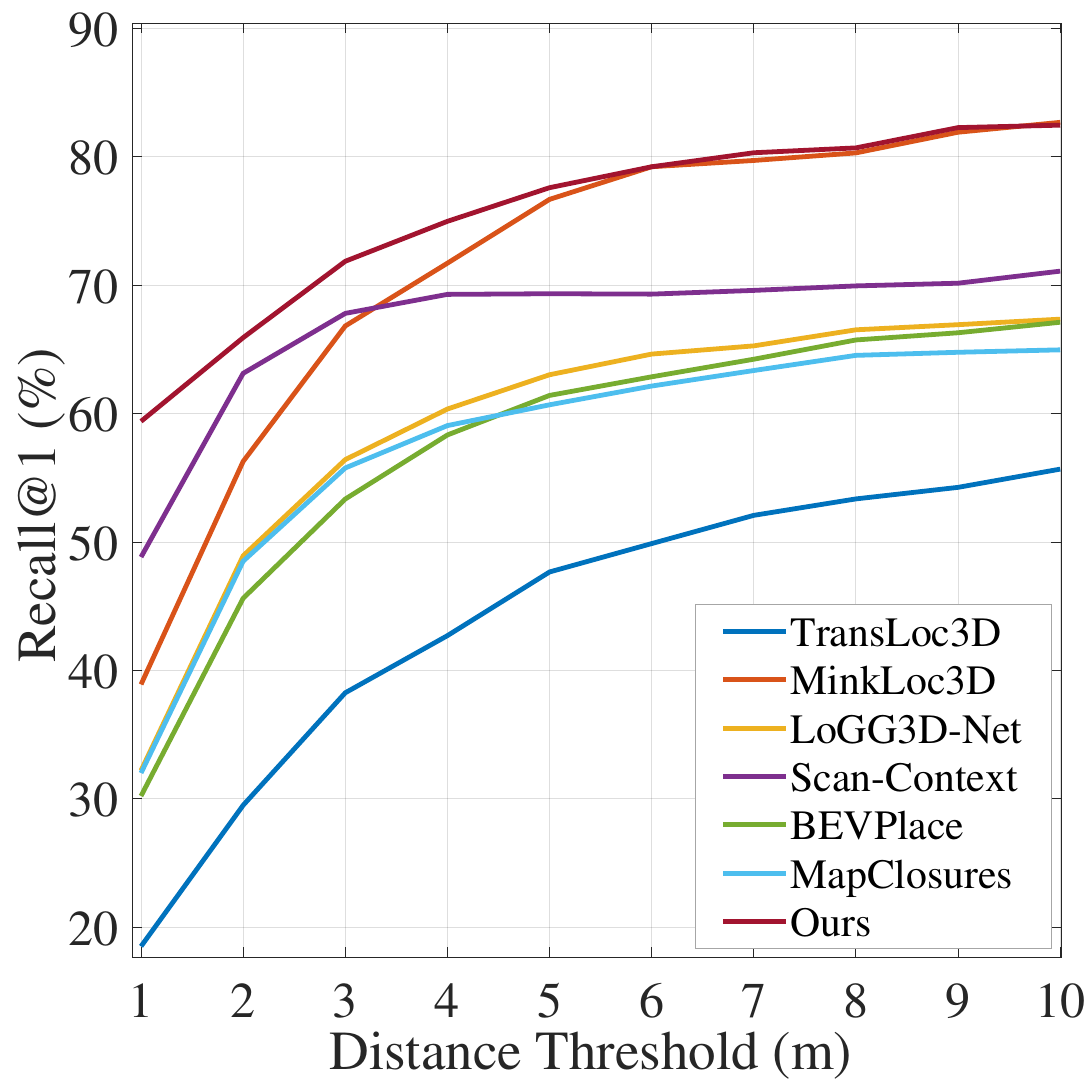}}
\subfloat[Karawatha-04]{
    \label{fig:k4}
  \includegraphics[width=0.48\columnwidth]{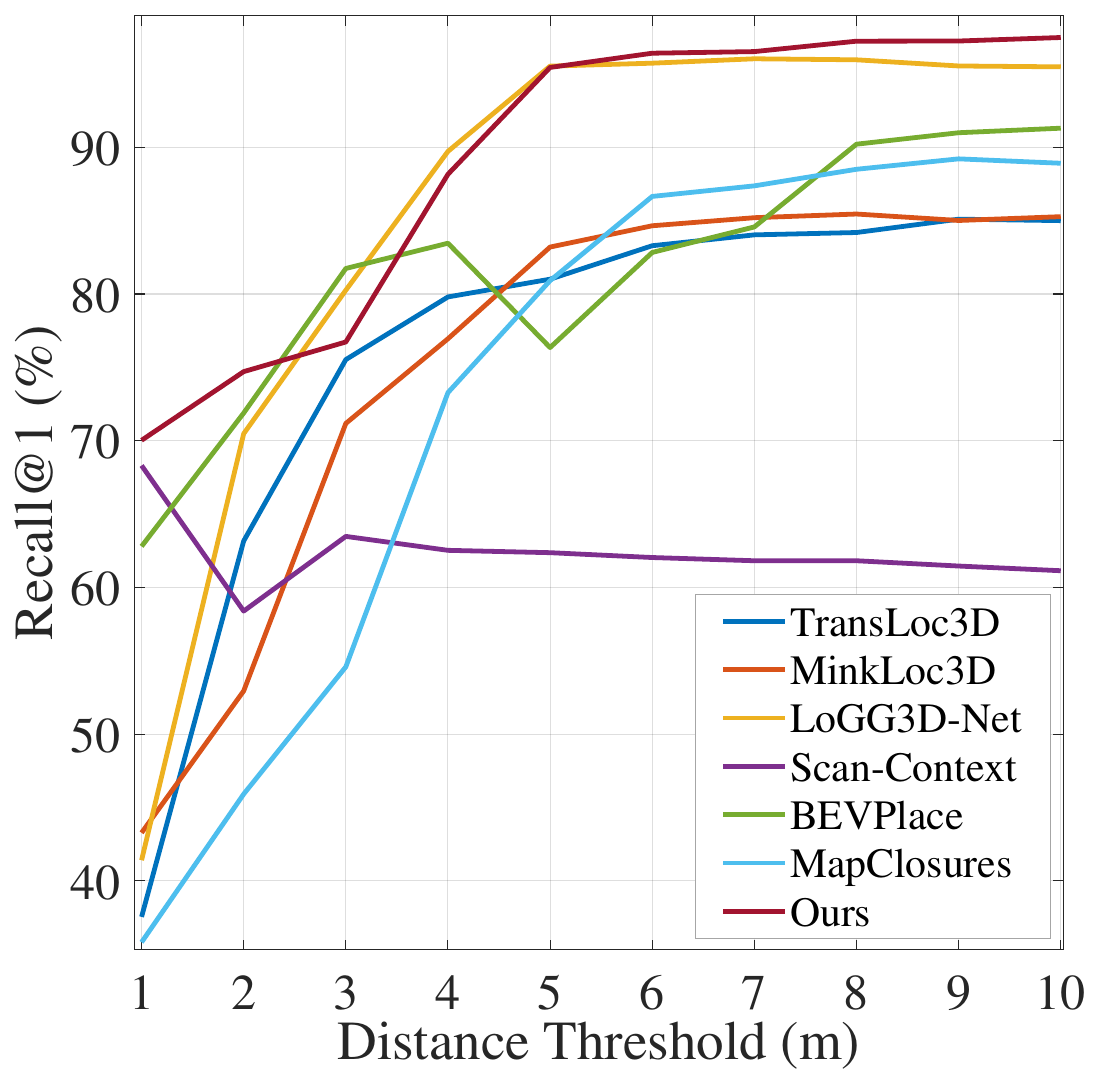}}
    \vspace{-0.2cm}
    \caption{{Recall@1 performance of methods for intra-sequence evaluations under different distance threshold. When calculating R@1, the denominator is the number of queries with corresponding positive samples, and the numerator is the number of queries that can retrieve positives. As the threshold increases, corresponding positives of one query may emerge, increasing the denominator by one and affecting monotonicity. The submap radius is \SI{30}{\meter}.}} 
    \label{fig:pair}
\end{figure}

\subsection{Qualitative Results}
\label{sec:visual}

During evaluation, we use \SI{3}{\meter} as the threshold to distinguish positives from the database.
Here, we give the R@1 performance curves of methods under different distance thresholds in \fref{fig:pair}.
ForestLPR is the most accurate method at \SI{1}{\meter} threshold, for fine localization tasks.
It can also be shown that as the distance threshold increases, only ForestLPR and MinkLoc3Dv2 achieve more than 80\% R@1 on the ANYmal dataset, and only ForestLPR and LoGG3D-Net achieve more than 95\% R@1 on the K-04 dataset with a \SI{10}{\meter} threshold, for other applications where a coarser localization suffices. 
Since the K-04 dataset is large and its retrieval space is also large, some curves in \fref{fig:k4} fluctuate, and scan-context even worsens as the threshold increases.
This means that it has random matches, resulting in the proportion decline.
The rising curves illustrate that most so-called failure cases are in fact coming from nearby places (but beyond the threshold) and not really random matches.

\begin{figure}[t]
    \centering
    \includegraphics[width=0.9\linewidth]{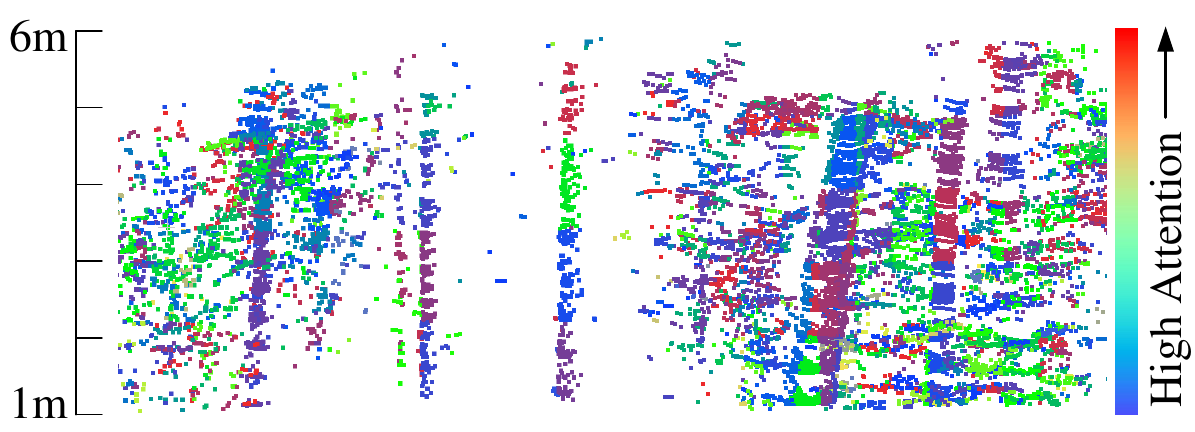}
    \vspace{-0.2cm}
    \caption{{Front-view visualization of the point cloud with projected patch-level attention. It can be seen that different heights are given different weights, especially the trees in the middle.}}
    \label{fig:weight}
\end{figure}

As shown in \fref{fig:weight}, our multi-BEV interaction module can adaptively pay attention to different heights at each patch location rather than relying only on any specific slice.
The regions with the highest weights predominantly correspond to areas least affected by seasonal changes.
This proves the effectiveness of our proposed interaction module and verifies our hypothesis that the geometric spatial distribution of trees and the discriminative features for different trees can be found at different heights.

\section{Discussion and Conclusion}
This work presents ForestLPR, a LiDAR-based place recognition method using multiple BEV density images in natural forests.
The proposed approach uses a transformer-based attention mechanism to capture contextual features, which utilize the spatial distribution information of trees.
Moreover, by generating multiple BEV images at different heights and introducing a multi-BEV interaction module, the proposed method can adaptively select the distinguishable features at different heights and effectively handle complex tree growth scenarios.
Remarkably, it outperforms previous methods on most major forest datasets and shows generalization on datasets with different forest environments and scan patterns.
Given its low computational cost, our method also has the potential to be deployed onboard robotic missions.
Since our method is proposed for forests and has only been tested on forest-like datasets, it is still uncertain about its generalization to dense jungles, where it is likely that all methods would have difficulties.

\section*{Acknowledgments}
This work was partly supported by the National Natural Science Foundation of China under Grants 62088102. Additionally, this work was partially supported by the EU Horizon 2020 programme grant agreement No. 101070405, the ETH Zurich Research Grant No. 21-1 ETH-27 and the Swiss National Science Foundation (SNSF) through project No. 227617.
{Thanks to Olga, Vaishakh, Andrei, and all my friends from ETHz RSL.}

{
    \small
    \bibliographystyle{ieeenat_fullname}
    \bibliography{ref}
}


\end{document}


\title{ForestLPR: LiDAR Place Recognition in {Forests} Attentioning Multiple BEV Density Images\\Supplementary Material}

\maketitle

This supplementary material is structured as follows.
In \sref{sec:dataset}, we provide more details about the datasets and evaluation metrics.
\sref{sec:imple} provides the implementation details of our method and the compared baselines.
In \sref{sec:ablation}, we show some additional ablation studies.
Finally, \sref{sec:result} contains additional quantitative results and qualitative results.

\section{Dataset and Evaluation Metrics Details}
\label{sec:dataset}
\textbf{Wild-Places.}
For both environments, Sequences 01 and 02 were collected on the same day, with Sequence 02 following the reverse route of Sequence 01. 
Sequence 03 was collected six months later and followed extended alternative routes, while Sequence 04 was collected 14 months after Sequence 01 and followed the same routes.
Sequences 03 and 04 are reserved for intra-sequence loop closure detection, and all sequences are used for inter-sequence evaluation to test the challenges posed by long-term variations.

Along with accurate 6DoF ground truth, each submap corresponds to a .pcd file containing the x, y, and z values of its points. All submaps are sampled from the map within a one-second window at the corresponding timestamp.
For training and validation, to more objectively evaluate the algorithm capabilities, we exclude the query neighboring frames that are temporally adjacent to the query from positive samples. This processing is similar to intra-sequence validation.

\begin{figure}[t]
    \centering
    \subfloat[Wild-Places]{\includegraphics[width=0.47\linewidth]{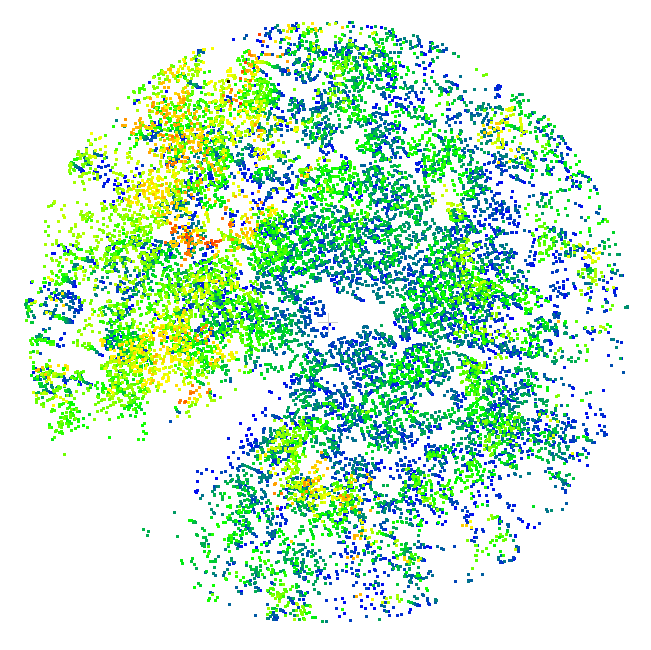}} \quad
    \subfloat[ANYmal Dataset]{\includegraphics[width=0.47\linewidth]{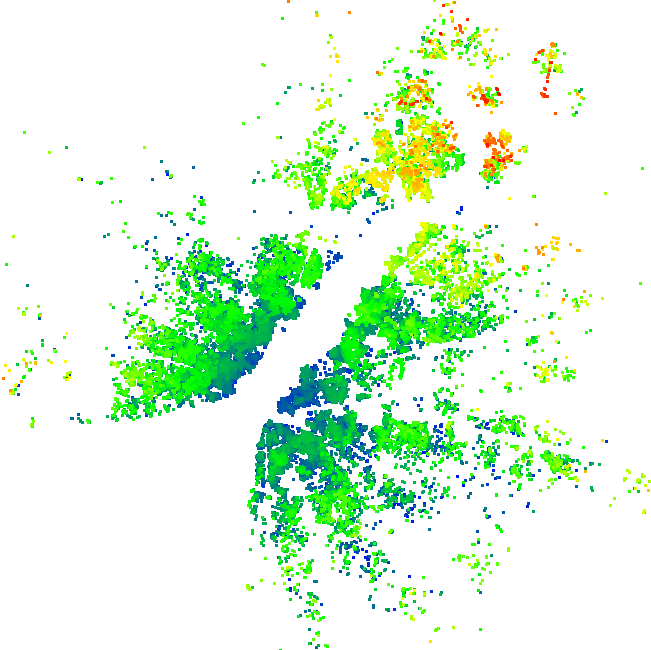}}
    \caption{BEV visualization of point clouds in Wild-Places and ANYmal dataset. The points are colored by the height: red means the highest one, and blue means the lowest one. There are blind spots in the lower left area.}
    \label{fig:blind}
\end{figure}

\textbf{ANYmal Dataset.}
The dataset was gathered in forests by following a triangular route twice and once counterclockwise.
Specifically, we first use Open3D SLAM~\cite{jelavic2022open3d} to generate the global poses and trajectory and remove frames with no or slight motion.
For our collected 10 HZ scan data, every five frames are selected as keyframes, and the distance between them is about \SI{0.5}{\meter}.
We also only sample points within a one second window of the corresponding timestamp for the submap.
This means that a sequence of consecutive scans $\{ \mathcal{P}_0, ..., \mathcal{P}_n \}$ are accumulated and transferred into the frame of middle scan $\mathcal{P}_{n/2}$, i.e., the keyframe.
Finally, only the points with a diameter of \SI{60}{\meter} are preserved.
Compared to Wild-Places, the ANYmal dataset was collected by a quadruped robot with a lower LiDAR viewpoint, leading to greater LiDAR angular variations during robot walking.
Furthermore, the LiDAR sensor is installed at the front-top of the robot, leading to a large blind area in scans (see \fref{fig:blind}).
The dataset contains a total of 1849 submaps (multi-frame registered scan samples) and 1239 loop closure-revisit pairs that meet the threshold, half of which are reverse revisits.

\textbf{Botanic Dataset.}
We repurpose the \textit{BotanicGarden} dataset~\cite{liu2023botanicgarden}, which is proposed for robot navigation in unstructured natural environments.
Its submap generation is identical to the ANYmal dataset.


\begin{table}[t]
    \centering
    \caption{Platform comparison.}
    \label{tab:data}
    \vspace{-0.3cm}
    \scalebox{0.9}{\begin{tabular}{cp{2cm}p{3.5cm}}
    \toprule
        Dataset & Platform & LiDAR mount condition\\
    \midrule
        Wild-Places & handheld sensor payload & an angle of 45°, \SI{1.5}{\meter} above the ground \\
        ANYmal & quadrupedal robot & vertically, \SI{0.7}{\meter} above the ground \\
        Botanic & wheeled robot & vertically, \SI{1.16}{\meter} above the ground \\
    \bottomrule
    \end{tabular}}
\end{table}

Here we summarize the LiDAR mounting configurations (e.g., positions and angles).
Wild-Places is collected by a handheld sensor payload, which includes a LiDAR mounted at an angle of 45°, \SI{1.5}{\meter} above the ground. 
ANYmal dataset is collected by the quadrupedal robot, which contains a vertically mounted VLP-16, {\SI{0.7}{\meter}} above the ground.
Botanic dataset is collected by a wheeled robot Scout V1.0 from AgileX, which contains a vertically mounted LiDAR, \SI{1.164}{\meter} above the ground.

\textbf{Evaluation Metrics.}
Following the setting in~\cite{wildplaces}, we consider a query successfully localized if the retrieved candidate is within \SI{3}{\meter}.

For intra-sequence evaluation on all datasets, we use Recall@1 (R@1) and maximum F1 score (F1) as the metric.
In particular, a query image is considered correctly localized if at least one of the top $N$ ranked reference images is a positive candidate.
In addition, previous entries adjacent to the query by less than $t$ time difference are excluded from the search to avoid matching to the same/nearby instances. For Wild-Places, $t=600$. For other datasets with a shorter sequence length, $t=100$.

\begin{table*}[t]
  \centering
   \caption{Ablation studies on BEV generation. ``elevation'' denotes using maximum height to generate BEV images, and the results with ``density'' are the same as those in Table 1 and Table 2 of the main paper.}
  \tabcolsep=0.17cm
  \label{tab:abaltion_height}
   \scalebox{1.0} {
  \begin{tabular}{cccccccccccccc}
  \toprule
{\multirow{2}{*}{Method}}& \multirow{2}{*}{Setting}&\multicolumn{2}{c}{V-03} & \multicolumn{2}{c}{V-04} & \multicolumn{2}{c}{K-03} & \multicolumn{2}{c}{K-04}& \multicolumn{2}{c}{Inter-V} & \multicolumn{2}{c}{Inter-K}\\
\cmidrule(lr){3-4} \cmidrule(lr){5-6} \cmidrule(lr){7-8} \cmidrule(lr){9-10}  \cmidrule(lr){11-12} \cmidrule(lr){13-14}
&& F1 & R@1 & F1& R@1 & F1 & R@1 & F1 & R@1  & R@1 & MRR & R@1 & MRR\\
\midrule
\multirow{2}{*}{\makecell{Scan \\ Context~\cite{scan}}}
&elevation&8.54&34.73&23.28&50.05&15.40&{35.03}&42.56&{55.91}&36.37&41.39&{24.88}&{30.44}\\  
&density&37.66&54.80&64.49&75.99&34.33&50.23&66.85&{63.48}&57.23&58.11&52.81&55.82\\  
\midrule
\multirow{2}{*}{\makecell{Single BEV}}
& elevation&42.35&46.64&32.80&43.01&39.42&51.56&54.36&{61.93}&34.20&55.84&35.07&56.73\\
& density&52.17&53.70&68.07&64.46&58.85&54.72&67.93&70.03&52.59&69.87&53.45&69.71\\
\midrule
\multirow{2}{*}{\makecell{Ours}}
& elevation&44.81&63.09&39.20&58.47&43.40&59.51&56.77&68.08&45.76&64.34&46.81&65.78\\
& density&{64.15}&{76.53}&{78.62}&{82.33}&{65.01}&{74.89}&{81.97}&{76.73}&{77.14}&{84.26}&79.02&83.87\\
\bottomrule
\end{tabular}}
\end{table*}

For inter-sequence evaluation on Wild-Places, Recall@1 and mean reciprocal rank (MRR) are used, where 
$$MRR = \frac{1}{N} {\textstyle \sum_{i=1}^{N}} \frac{1}{rank_i},$$ 
and $rank_i$ is the rank of the first retrieved positive candidate to query submap $i$.
If none of the top 25 candidates are correct, the reciprocal rank is 0.
The final R@1 and MRR values are the means of the respective R@1 and MRR values overall evaluations.

\section{More Implementation Details}
\label{sec:imple}

PointPillar~\cite{lang2019pointpillars}, TransLoc3D~\cite{transloc3d}, MinkLoc3Dv2~\cite{minkloc3dv2}, LoGG3D-Net~\cite{logg3d}, and BEVPlace~\cite{luo2023bevplace} are all state-of-the-art learning-based methods and have saturated performance on the popular urban LiDAR datasets. 
ScanContext~\cite{scan} and MapClosures~\cite{mapclosure} are handcrafted approaches, which first encode the highest z-value of bins or extract FAST and ORB features and use the key points for loop detection.

{The implementation details of all compared baselines in Table 1 and 2 of the main paper are as follows:}
\begin{itemize}
    \item PointPillar\footnote{\href{https://github.com/zhulf0804/PointPillars}{github.com/zhulf0804/PointPillars}}: Remove the classifcation head and add another GeM layer and L2Norm to generate the final features. Use pre-processed point cloud as input and Triplet loss to finetune the backbone.
    \item TransLoc3D\footnote{\href{https://github.com/slothfulxtx/TransLoc3D}{github.com/slothfulxtx/TransLoc3D}}: Use \textit{configs/transloc3d\_baseline\_cfg.py} and set quantization\_size to 1.
    \item MinkLoc3Dv2\footnote{\href{https://github.com/jac99/MinkLoc3Dv2}{github.com/jac99/MinkLoc3Dv2}}: Use \textit{models/minkloc3dv2.txt}. Set coordinates to polar, quantization\_step to (0.8, 0.15, 0.15), and normalize\_embeddings to True.
    \item LoGG3D-Net\footnote{\href{https://github.com/csiro-robotics/Wild-Places/tree/main/scripts/eval/logg3d}{github.com/csiro-robotics/Wild-Places/scripts/eval/logg3d}}: Use the default setting in Wild-Places.
    \item ScanContext\footnote{\href{https://github.com/csiro-robotics/Wild-Places/tree/main/scripts/eval/scancontext}{github.com/csiro-robotics/Wild-Places/scripts/eval/scancontext}}: Use the default setting. To compare feature extraction modules fairly, we change the elevation BEV images generated from the whole point cloud to \textit{density BEV images generated from pre-processed point cloud}, which is the same as ours.
    \item BEVPlace\footnote{\href{https://github.com/zjuluolun/BEVPlace}{github.com/zjuluolun/BEVPlace}}: We use adjusted BEV density images as input and finetune the model with triplet loss, margin = 0.3. 
    \item MapClosures\footnote{\href{https://github.com/PRBonn/MapClosures}{github.com/PRBonn/MapClosures}}: Similarly, we use the adjusted BEV density images as input. Following the setting in the paper, we use a threshold of 50 bits on the Hamming distance for descriptor match. The threshold for the number of inliers obtained from RANSAC alignment is set to 10.
    \item Ours hyper-parameter selection: \textit{1}. Multi-level feature extraction: we refer to the settings in~\cite{transvpr}.
\textit{2}. Based on the distance threshold (\SI{3}{\meter}), we adjusted the overlap threshold for comparability.
\textit{3}. Backbone design: we adopted commonly used hyper-parameters in computer vision.
\end{itemize}

\begin{table*}[t]
  \centering
   \caption{Ablation studies on slice number when generating multiple BEV density images from pre-processed point clouds. 
   * represents the results already given in our main paper.}
  \tabcolsep=0.12cm
  \label{tab:abaltion_hyper}
   \scalebox{0.84} {
  \begin{tabular}{cccccccccccccccccccc}
  \toprule
\multirow{2}{*}{\makecell{Slice \\ Number }}& \multirow{2}{*}{$\bigtriangleup h$}&\multicolumn{2}{c}{V-03} & \multicolumn{2}{c}{V-04} & \multicolumn{2}{c}{K-03} & \multicolumn{2}{c}{K-04}& \multicolumn{2}{c}{Inter-V} & \multicolumn{2}{c}{Inter-K} & \multicolumn{2}{c}{ANYmal} & \multicolumn{2}{c}{Botanic-03} & \multicolumn{2}{c}{Botanic-06}\\
\cmidrule(lr){3-4} \cmidrule(lr){5-6} \cmidrule(lr){7-8} \cmidrule(lr){9-10}  \cmidrule(lr){11-12} \cmidrule(lr){13-14} \cmidrule(lr){15-16} \cmidrule(lr){17-18} \cmidrule(lr){19-20}
&& F1 & R@1 & F1& R@1 & F1 & R@1 & F1 & R@1  & R@1 & MRR & R@1 & MRR & F1 & R@1 & F1 & R@1 & F1 & R@1\\
\midrule
\midrule

10 & 0.5& 57.63&70.81&73.15&77.83&60.35&68.12&77.48&70.54&71.59&76.34&73.86&74.19&75.02&70.43&71.87&79.36&73.07&76.21\\
5 & 1* &{64.15}&{76.53}&{78.62}&{82.33}&{65.01}&{74.89}&{81.97}&{76.73} &{77.14}&{84.26}&{79.02}&{83.87}&{81.45}&{71.87}&{78.21}&{84.81}&{78.82}& {82.00}\\
2 & 2.5& 55.68&67.91&70.86&75.19&58.71&63.78&72.57&65.93&64.71&72.48&70.36&71.25&74.31&70.82&70.34&77.56&71.94&75.92\\
\bottomrule
\end{tabular}}
\end{table*}

\begin{table*}[t]
  \centering
   \caption{Ablation studies on positive sample mining strategy. All ablations use our final backbone and are trained on Wild-Places.}
  \tabcolsep=0.12cm
  \label{tab:abaltion_strategy}
   \scalebox{0.835} {
  \begin{tabular}{ccccccccccccccccccc}
  \toprule
\multirow{2}{*}{Strategy}&\multicolumn{2}{c}{V-03} & \multicolumn{2}{c}{V-04} & \multicolumn{2}{c}{K-03} & \multicolumn{2}{c}{K-04}& \multicolumn{2}{c}{Inter-V} & \multicolumn{2}{c}{Inter-K} & \multicolumn{2}{c}{ANYmal} & \multicolumn{2}{c}{Bo.-03} & \multicolumn{2}{c}{Bo.-06}\\
\cmidrule(lr){2-3} \cmidrule(lr){4-5} \cmidrule(lr){6-7} \cmidrule(lr){8-9}  \cmidrule(lr){10-11} \cmidrule(lr){12-13} \cmidrule(lr){14-15} \cmidrule(lr){16-17} \cmidrule(lr){18-19}
& F1 & R@1 & F1& R@1 & F1 & R@1 & F1 & R@1  & R@1 & MRR & R@1 & MRR & F1 & R@1 & F1 & R@1 & F1 & R@1\\
\midrule
Distance&57.58&70.14&75.86&78.12&60.95&70.81&78.34&72.98& 72.56&76.85&75.41&76.48&76.46&67.59&74.31&80.86&75.49&79.63 \\
Overlap\cite{chen2021overlapnet,mapclosure}&61.04 & 74.82&77.91&80.26&63.25&73.59&80.31&74.92&75.08&78.31&77.63&78.52&80.03&70.51&75.96&81.98&77.38&80.14\\
Ours&{64.15}&{76.53}&78.62&{82.33}&{65.01}&{74.89}&{81.97}&{76.73} &77.14&84.26&79.02&83.87&{81.45}&{71.87}&{78.21}&{84.81}&{78.82}& {82.00} \\
\bottomrule
\end{tabular}}
\end{table*}

\section{Additional Ablation Studies}
\label{sec:ablation}
First, we give more results to show the difference between density or height when generating BEV images for our model and Scan-Context.
Then, we evaluate the model with different hyper-parameters.
Finally, we present evaluation results for the model trained with a distance-based sample mining strategy and other similar overlap-based strategies \cite{mapclosure,chen2021overlapnet}.

\textbf{BEV Images Generation.}
We ablate our choice of density images over elevation images by modifying our pipeline and Scan-Context in intra- and inter-sequence evaluations of Wild-Places.
Specifically, we use pre-processed point clouds as input and generate elevation images.

As shown in \tref{tab:abaltion_height}, for all three models, although BEV elevation images can also reflect the spatial distribution of trees, density is better than elevation in all tests.
On the one hand, the pre-processed point cloud has removed points above  \SI{6}{\meter}, and the same elevation of many pixels causes the loss of features. 
On the other hand, the elevation image is also sensitive to the orientation of the sensor (pitch and roll), as the maximum height recorded varies with the distance and occlusions between the scanner and the object.
{More qualitative results refers to \sref{sec:visual}.}


\textbf{Hyper-parameters Sensitivity.}
Considering that $S$ (the number of slices) and $\bigtriangleup h$ (height interval of slices) setting in height cropping are the most critical hyper-parameters for our work, we perform ablation experiments on them by setting $\bigtriangleup h=\SI{0.5}{\meter}, \SI{2.5}{\meter}$ respectively.

For setting different $\bigtriangleup h$, the results show that \SI{1}{\meter} is a better choice. When digging into the samples, we find that 0.5m is too small to obtain effective density information and 2m is too large, causing confusion between different heights.
So, we select \SI{1}{\meter} as the final hyper-parameter in our main paper based on experience.





\textbf{Positive Sample Mining Strategy.}
In this experiment, we adopt the same global features from our model and compare our method based on different ground truth definitions (i.e., positive sample mining strategies).
``Distance'' uses the ground truth based on physical distance 12.5m as the threshold for positive examples and 50m as the threshold for negative examples), and ``Overlap'' denotes the similar overlap-based method~\cite{chen2021overlapnet,mapclosure} (0.9 as the threshold for positives and 0.5 as the threshold for negatives), which is similar to \cite{leyva2023data} for visual place recognition.

\begin{figure}[t]
    \centering
\subfloat[Reverse visits with same physical distance.]{
  \includegraphics[width=0.9\linewidth]{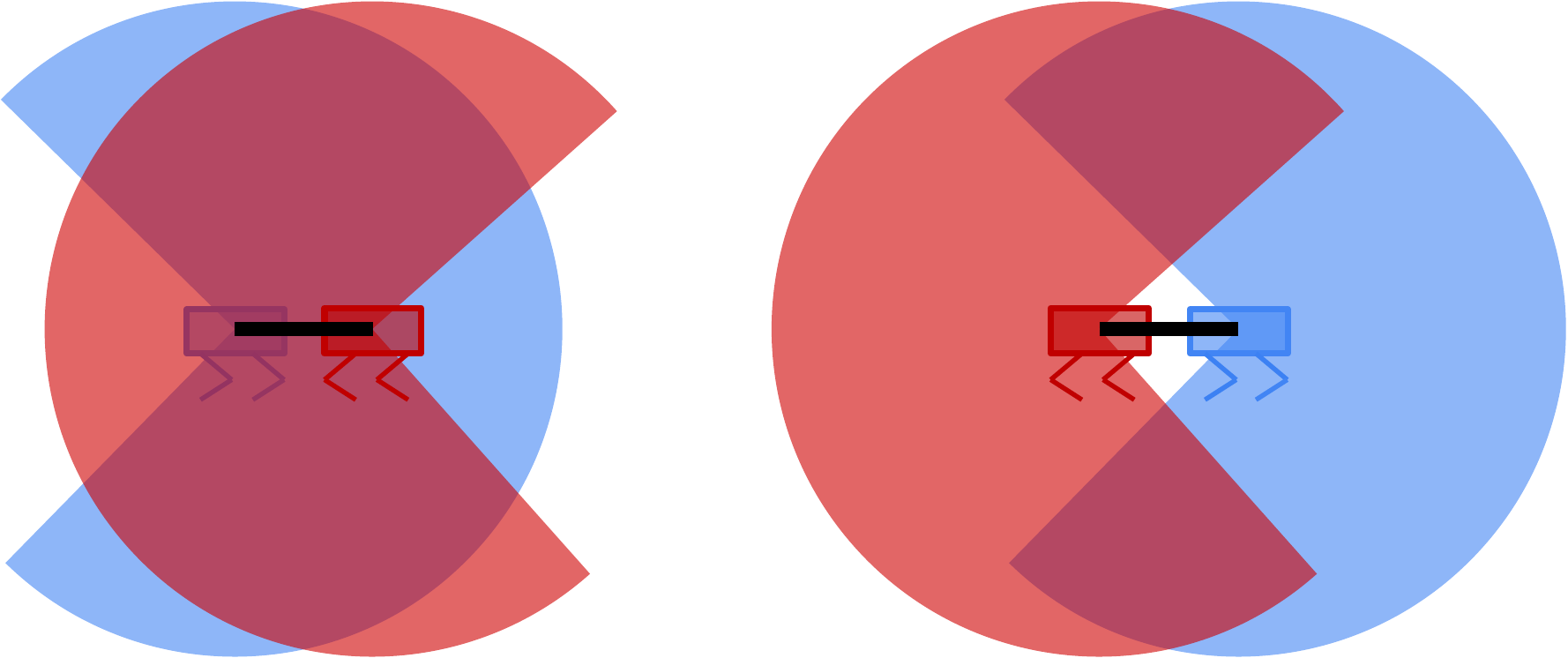}}\\
\subfloat[Visits with same physical distance.]{
  \includegraphics[width=0.9\linewidth]{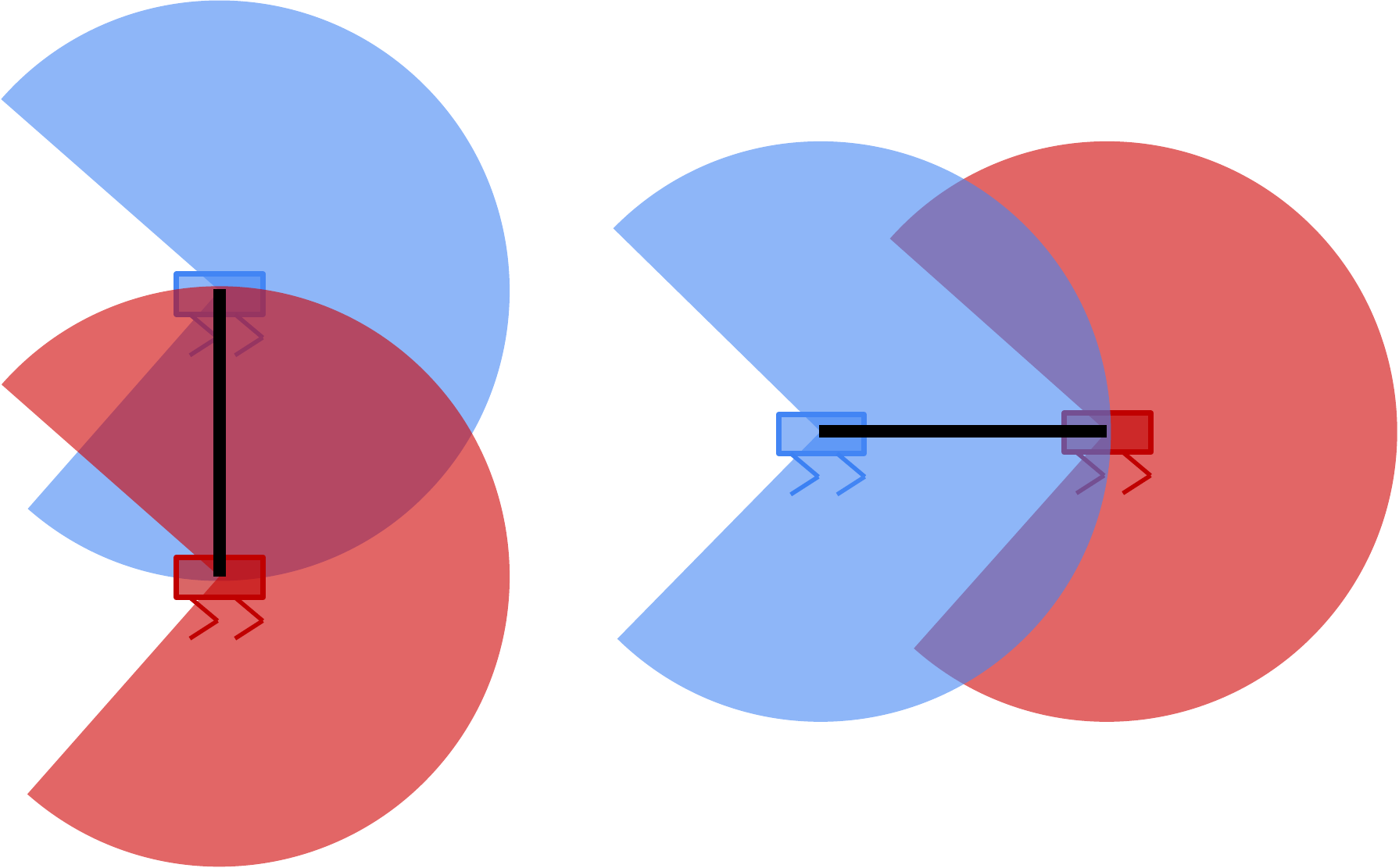}}
    \caption{Illustration of the effect of blind spots on the degree of common vision between two frames of point clouds.}
    \label{fig:overlap}
\end{figure}

\begin{table*}[t]
  \centering
   \caption{Additional studies on removing ground and tree top from point clouds. Here we show the results of the methods with cropping (pre-processed point clouds) and without cropping (raw point clouds).}
  \tabcolsep=0.2cm
  \label{tab:abaltion_crop}
   \scalebox{0.95} {
  \begin{tabular}{lccccccccccccc}
  \toprule
{\multirow{2}{*}{Method}}& {\multirow{2}{*}{Crop}} &\multicolumn{2}{c}{V-03} & \multicolumn{2}{c}{V-04} & \multicolumn{2}{c}{K-03} & \multicolumn{2}{c}{K-04} & \multicolumn{2}{c}{inter-V}& \multicolumn{2}{c}{inter-K}\\
\cmidrule(lr){3-4} \cmidrule(lr){5-6} \cmidrule(lr){7-8} \cmidrule(lr){9-10}  \cmidrule(lr){11-12} \cmidrule(lr){13-14}
&& F1 & R@1 & F1& R@1 & F1 & R@1 & F1 & R@1 & R@1 & MRR & R@1 & MRR\\
\midrule
{\multirow{2}{*}{Scan-Context~\cite{scan}}}
&× &4.77&11.03&36.19&43.01&30.65&45.81&50.58&60.33&46.76&48.87&\underline{56.40}&\underline{59.57}\\  
&\checkmark&37.66&54.80&64.49&75.99&34.33&50.23&66.85&63.48&57.23&58.11&\underline{52.81}&\underline{55.82}\\ 
{\multirow{2}{*}{BEVPlace~\cite{luo2023bevplace}}}
& × &0.73&2.32&28.65&48.13&26.83&{34.68}&60.59&{73.61}&33.93&56.16&34.99&58.51\\ 
& \checkmark&6.79&24.04&43.67&63.50&32.87&40.62&60.82&81.74&51.91&67.68&41.40&61.59\\ 
{\multirow{2}{*}{MapClosures~\cite{mapclosure}}}
& × &12.99&6.95&29.44&14.94&\underline{25.05}&\underline{14.35}&65.38&\underline{55.05}&26.68&27.44&\underline{19.48}&20.22\\
& \checkmark&38.47&23.82&49.95&25.61&\underline{21.35}&\underline{11.95}&70.62&\underline{54.59}&34.31&37.83&\underline{19.35}&21.16\\
\bottomrule
\end{tabular}}
\end{table*}

The results in \tref{tab:abaltion_strategy} indicate that overlap-based mining is more helpful than physical distance-based mining, especially for V-03, which contains a large number of reverse revisits. That's because the overlap calculation directly corresponds to the degree of common vision.
As shown in \fref{fig:overlap}, for data with blind spots, physical distance and degree of common vision are no longer proportional, and it's more reasonable to consider perspective simultaneously.
In addition, our overlap calculation is:
\begin{equation}
\label{eq:overlap}
    o\left(\mathcal{V}_q, \mathcal{V}_p\right)=\frac{\left|\mathcal{V}_q \cap \mathcal{V}_p\right|}{\left|\mathcal{V}_q \cup \mathcal{V}_p\right|},
\end{equation}
which is more suitable for handling blind spots than those works~\cite{chen2021overlapnet,mapclosure} that utilize $min(\cdot)$ as denominator.

In some extreme cases, the valid points within the scan range of one sample may completely cover another.
Then, the similarity calculated by~\cite{chen2021overlapnet,mapclosure} is 1, which is unreasonable, while \eref{eq:overlap} can better ensure the rationality of sample mining.

\section{Additional Results}
\label{sec:result}

\begin{figure*}[t]
    \centering
    \includegraphics[width=\linewidth]{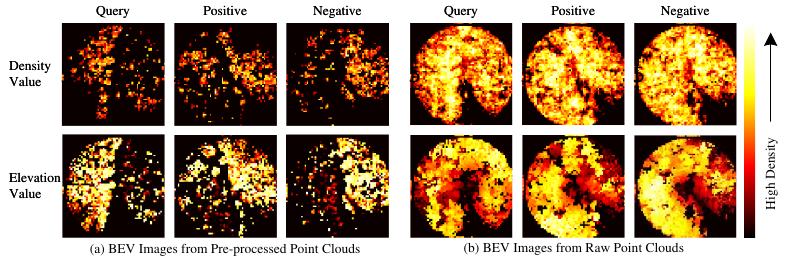}
    \caption{Generated single BEV images from point clouds. (a) shows the BEV images from pre-processed point clouds, and (b) shows ones from raw point clouds. It can be shown that the single BEV density image generated from pre-processed point clouds contains clearer spatial information than others, although it is still insufficient for place recognition in forests.}
    \label{fig:bev}
\end{figure*}

\subsection{Point Cloud Cropping}
As mentioned in the main paper, the results of BEV-based methods in Tables 1 and 2 are also computed from pre-processed point clouds.

To prove that ground and tree top removal is general for single BEV-based baselines in forests, we give more results in \tref{tab:abaltion_crop}.
For single BEV-based baselines without cropping, we use whole point clouds to generate BEV density images and re-train BEVPlace on Wild-Places.
It is worth noting that we have tried using pre-processed point clouds as input and training LoGG3D-Net on Wild-Places, but the performance was poor.
We speculate that it is a problem with the hyper-parameter settings during training. For fairness, we do not give the results here for comparative analysis.

As shown in \tref{tab:abaltion_crop}, using the whole point cloud has a detrimental impact on all single BEV-based methods, with performance drops by up to more than 40\% across the board.
It demonstrates the importance of removing ground and tree tops for single BEV-based place recognition methods in forests.

As the special case in \tref{tab:abaltion_crop}, Scan-Context without cropping achieves better performance on inter-K than the cropped one. 
MapClosures also shows some opposite situations on K-03/04 and inter-K.
Both of these two methods are non-learning based.
This may be because trees in Karawatha are not as tall as those in Venman, and more discriminative density information from the tree top can be captured during scanning. (more visualization examples are shown in \ref{sec:visual}.)

In a word, the results provide solid insights about removing ground and tree tops in forests to guide future related works.

\subsection{Qualitative Results}
\label{sec:visual}

\begin{figure*}[t]
    \centering
    \subfloat[Height histogram of a submap in K-03]{\includegraphics[width=0.5\linewidth]{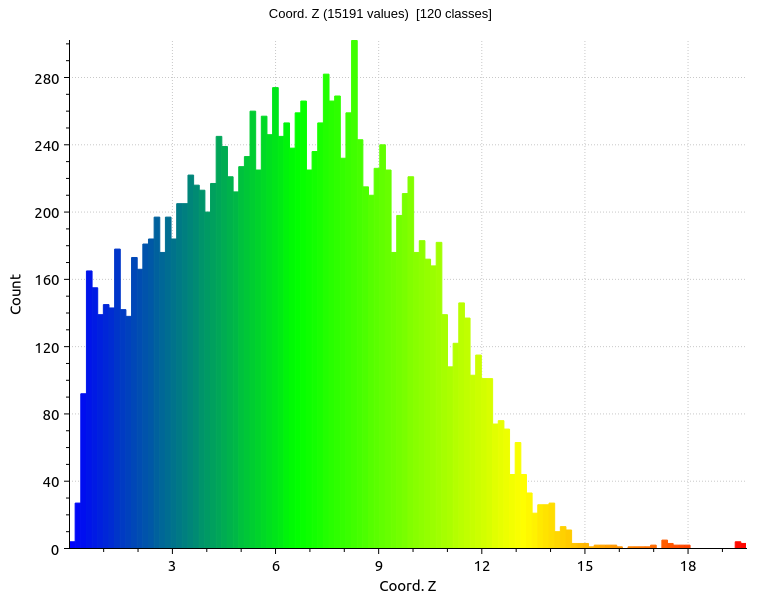}} 
    \subfloat[Height histogram of a submap in V-03]{\includegraphics[width=0.5\linewidth]{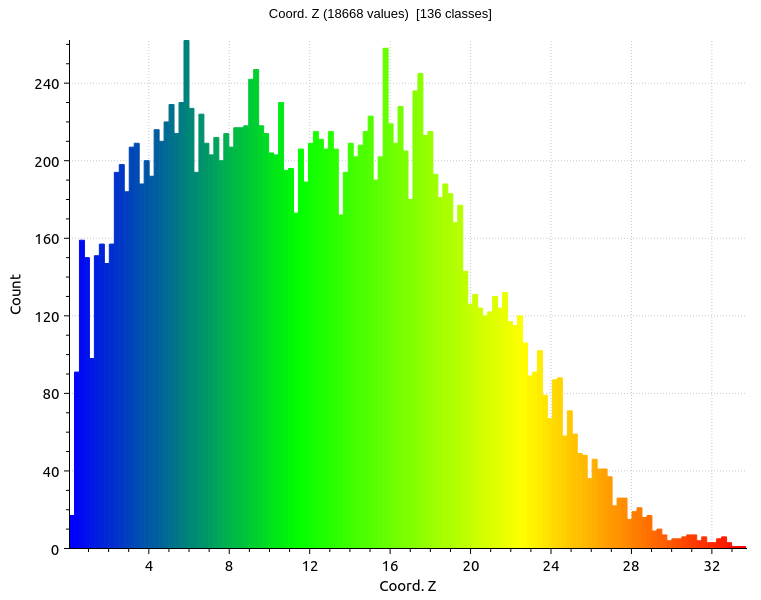}}
    \caption{Height histogram of submaps in Karawatha-03 and Venman-03 sequences. It can be seen that the growth of vegetation in K-03 is shorter than that in V-03.}
    \label{fig:height}
\end{figure*}

As shown in~\fref{fig:bev}, BEV images from the cropped point cloud are more recognizable, and the ones using the density value are easier to identify.
The BEV elevation images are sensitive to poses because the coordinate values of points can severely vary when the robot moves. In contrast, the BEV density images are more robust because point cloud density does not depend on specific points.

\fref{fig:height} gives the height histograms of random samples in the Karawatha-03 and Venman-03 sequence, illustrating the height difference of trees in the Karawatha and Venman datasets.
It supports our speculations that some algorithms present different performance levels on different sequences due to different scan patterns.

\subsection{Inter-sequence Evaluation}

As a supplement to TABLE 2 in the main paper, we provide the complete results of all compared methods on inter-sequence evaluations in \tref{tab:compare_SOTA_inter}, including R@1 and MRR scores.

\begin{table}[t]
  \centering
   \caption{Comparison with SOTA methods on Wild-Places. Validation refers to the testing dataset in training, and Inter-Venman/ Karawatha refers to Inter-sequence evaluations. Point-based methods are evaluated on raw point clouds. PointPillar and BEV-based methods are marked with *, showing the finetuning results on pre-processed point clouds and density images for fair comparison.}
  \label{tab:compare_SOTA_inter}
\scalebox{0.84}{
  \begin{tabular}{lcccccc}
  \toprule
{\multirow{2}{*}{Method}}& {Validation} & \multicolumn{2}{c}{Inter-V} & \multicolumn{2}{c}{Inter-K}\\
\cmidrule(l{-2pt}r{5pt}){2-2} \cmidrule(l{-2pt}r{5pt}){3-4} \cmidrule(l{0pt}r{0pt}){5-6} 
& R@1 & R@1 & MRR & R@1 & MRR  \\
\midrule
PointPillar*~\cite{lang2019pointpillars}&35.49&60.34&70.68&52.19&66.72\\
TransLoc3D~\cite{transloc3d} & 37.75&62.85&74.92&54.32&69.08\\
MinkLoc3Dv2~\cite{minkloc3dv2}& 51.40&75.77&\underline{84.87}&67.82&79.21\\
LoGG3D-Net~\cite{logg3d}& \underline{54.34}&\textbf{79.84} &\textbf{87.33} & \underline{74.67} &\underline{83.68}\\
Scan-Context*~\cite{scan}&-&57.23&58.11&52.81&55.82\\
BEVPlace*~\cite{luo2023bevplace}& 26.25&51.91&67.68&41.40&61.59\\
MapClosures*~\cite{mapclosure}&-&34.31&37.83&19.35&21.16\\
\midrule
Ours&\textbf{80.03}& \underline{77.14} & 84.26 &\textbf{79.02} & \textbf{83.87}\\
\bottomrule
\end{tabular}}
\end{table}


\bibliographystyle{ieeenat_fullname}
\bibliography{ref}